\documentclass[runningheads]{llncs}
\usepackage[T1]{fontenc}
\usepackage{graphicx}
\usepackage{booktabs}
\usepackage{multirow}
\usepackage{colortbl}
\usepackage{color}
\usepackage[pagebackref,breaklinks,colorlinks]{hyperref}

\usepackage{wrapfig}
\usepackage{orcidlink}
\usepackage{xr-hyper}
\usepackage{algorithm}
\usepackage{algpseudocode}
\usepackage{amsmath,bm}
\usepackage{bbm}
\usepackage{amsfonts}
\usepackage{caption}
\usepackage{subfig}
\usepackage[euler]{textgreek}
\usepackage{arydshln}
\usepackage{marvosym}

\definecolor{Gray}{gray}{0.9}

\def\rmX{{\mathbf{X}}}
\def\rmY{{\mathbf{y}}}
\def\rma{{\mathbf{a}}}
\def\rmt{{\mathbf{T}}}
\def\real{{\mathbb{R}}}
\def\rmW{{\mathbf{W}}}
\def\dlilp{\text{\tiny{DLILP}}}
\def\clip{\text{\tiny{CLIP}}}
\def\unicl{\text{\tiny{UniCL}}}
\def\uni{\text{\tiny{Uni}}}
\def\il{\text{\tiny{I-L}}}
\def\it{\text{\tiny{I-T}}}
\def\textCustome{\text{\tiny{text}}}
\def\img{\text{\tiny{img}}}
\def\itot{\text{\tiny{i2t}}}
\def\ttoi{\text{\tiny{t2i}}}

\definecolor{darkblue}{rgb}{0.21,0.49,0.74}
\newcommand{\appensecref}[1]{{Appendix~\ref{#1}}}

\begin{document}

\title{A Reality Check of Vision-Language Pre-training in Radiology: Have We Progressed Using Text?} 
\titlerunning{A Reality Check of Vision-Language Pre-training in Radiology}
\author{Julio Silva-Rodríguez\textsuperscript{\Letter} \and Jose Dolz \and  Ismail {Ben Ayed}}
\authorrunning{J.~Silva-Rodríguez et al.}
\institute{ÉTS Montréal \\
\Letter {\tt \small \email{julio-jose.silva-rodriguez@etsmtl.ca}}
}

\maketitle         

\begin{abstract}

Vision-language pre-training has recently gained popularity as it allows learning rich feature representations using large-scale data sources. This paradigm has quickly made its way into the medical image analysis community. In particular, there is an impressive amount of recent literature developing vision-language models for radiology. However, the available medical datasets with image-text supervision are scarce, and medical concepts are fine-grained, involving expert knowledge that existing vision-language models struggle to encode. In this paper, we propose to take a prudent step back from the literature and revisit supervised, unimodal pre-training, using fine-grained labels instead. We conduct an extensive comparison demonstrating that unimodal pre-training is highly competitive and better suited to integrating heterogeneous data sources. Our results also question the potential of recent vision-language models for open-vocabulary generalization, which have been evaluated using optimistic experimental settings. Finally, we study novel alternatives to better integrate fine-grained labels and noisy text supervision. Code and weights are available: \url{https://github.com/jusiro/DLILP} .
  
  
\end{abstract}

\section{Introduction}
\label{sec:intro}

The recent advancements in deep learning have yielded remarkable outcomes to enhance computer-aided medical image analysis \cite{mediasurv}. Nevertheless, these have been classically hampered by the necessity of using large labeled datasets for training successful specific solutions, which may not generalize properly under domain drifts \cite{Finlayson2021}. Currently, there is a paradigm shift led by multimodal foundation models. Such visual understanding models are pre-trained for specific domains using large dataset assemblies and heterogeneous learning objectives. In this way, foundation models learn rich generalizable features that can be efficiently adapted to downstream tasks \cite{clap24}. These conditions are ideal for the widespread adoption of deep-learning solutions to clinical institutions, with limited data and computational resources \cite{Malwina2022,Moor2023,foundMed}. In particular, vision-language contrastive pre-training methods such as CLIP \cite{clip} have revolutionized the computer vision field. These approaches train a joint multimodal space, in which text and visual data representations are aligned. Using web-mined data, CLIP gathers a collection of 400M image-text pairs for pre-training and has shown impressive generalization capabilities when transferred on various downstream computer-vision tasks. Driven by CLIP's popularity, vision-language models are also paving the way for building strong medical foundation models in different application domains, such as histology \cite{PLIP}, retina \cite{FLAIR}, and radiology \cite{refers}. In particular, radiology, and more concretely, chest X-ray image understanding, has been an essential focus of this emergent literature since radiology text reports are the \textit{de facto} raw supervisory information easily accessible from medical clinical records. A myriad of recent vision-language models, such as Convirt \cite{convirt}, REFERS \cite{refers}, GlorIA \cite{gloria}, MedCLIP \cite{medclip}, medKLIP \cite{medklip}, and others \cite{ked,cxrclip,mgca,biovil}, attest this trend. Many of these recent works are published in the top vision conferences or prestigious journals, advocating a paradigm shift in radiology imaging interpretation, driven by contrastive image-text pre-training. However, as we show, the potential to leverage large transferable vision models through more classical approaches, such as unimodal pre-training, has been severely underestimated.

Relying on text supervision for vision pre-training in medical domains faces several challenges. First, available datasets are orders of magnitude smaller compared to natural images. For example, these models are mainly built upon the textual information available in MIMIC \cite{mimic}, which assembles solely 257K image-text pairs. Second, as discussed in \cite{biovil}, medical linguistics are highly specialized and contain domain-specific structures. These include negations (e.g. “\textit{there is no consolidation}”), expressions of uncertainty (e.g., “\textit{possibly progressing to pneumonia}"), spatial relations (e.g., \textit{"bilateral heterogeneous airplane opacities}"), hierarchical relationships (e.g., \textit{"infection}" $\rightarrow$ \textit{"pneumonia}"), or abbreviations. Although some efforts have been devoted to regularize the training to focus on this information \cite{biovil,mrm}, vision-language pre-training struggles to properly encode such expert knowledge. Indeed, this is not only the case of medical knowledge. As recent studies show, vision-language models might struggle to properly codify basic spatial information \cite{whatsupvlms} or fine-grained vision-text correspondences \cite{tang2023when}. Thus, in addition to text supervision, recent works \cite{medclip,medklip,ked} have proposed using label information for aligning better image and text representations. These labels are obtained through entity extraction NLP methods, such as CheXpert-labeler \cite{chexpert} or RadGraph \cite{radgraph}, and follow radiologist-designed rule-based algorithms able to encode text reports to concrete labels through expert knowledge — \textit{note that these labels do not require costly manual image annotation}. Indeed, before the wave of vision-language models, these labels represented the predominant supervision for training dataset-specific deep learning models for chest X-rays, and an important number of datasets (e.g., CheXpert \cite{chexbert}, NIH \cite{nih}, or PadChest \cite{padchest}) included primarily image-label information. Nonetheless, even though these datasets contain fine-grained labels, supervised pre-training is being surprisingly overlooked in the current literature, even as a baseline to measure actual progress in the field. Based on these observations, we present the following contributions:
\begin{enumerate}
    \item We challenge the status quo of current contrastive vision-language models (VLMs) for visual comprehension of chest X-rays (CXR), advocating for revisiting \textbf{supervised pre-training}. In particular, we focus on evaluating their zero- and few-shot transferability on a broad 7-task benchmark.
    \item We demonstrate (see \textit{Observation 1}) that such unimodal pre-training is a largely competitive solution, able to integrate larger heterogeneous sources.
    \item In addition, we offer a critical view of the current trends in evaluating the zero-shot capabilities of CXR VLMs to novel diseases (see \textit{Observations 2 and 3}). Concretely, we show that local unspecific findings drive textual disease prototypes, and VLMs fail to distinguish between overlapping conditions.
    \item Finally, we investigate approaches for effectively integrating labels and noisy textual information. Concretely, we propose a novel \textbf{D}isentangled \textbf{L}anguage-\textbf{I}mage-\textbf{L}abel \textbf{P}re-training, \textbf{DLILP}. Unlike existing strategies, DLILP offers a robust trade-off for zero-shot generalization to both known and novel and is scalable to combining image-label and image-text datasets. 
\end{enumerate}

\section{Related Work}
\label{sec:rw}

\noindent\textbf{\textit{Pre-training and adapting visual recognition models}.} Current computer vision applications are fueled by transferring rich pre-trained representations learned on large-scale datasets. Traditionally, pre-training has been driven by human-annotated data for a given set of heterogeneous categories, such as ImageNet \cite{imagenet}, via standard cross-entropy or supervised contrastive \cite{supcons} objectives. More recently, leveraging large-scale datasets with text supervision has gained increasing interest within the computer vision community. In particular, foundation models such as CLIP \cite{clip} or ALIGN \cite{align} have shown great success in zero-shot generalization and efficient transfer learning following multimodal contrastive learning. To also integrate discriminative, label-driven information, UniCL \cite{unicl} proposed a unified framework by aligning image, text, and label spaces into the same optimization criteria. While UniCL showed superior performance to its supervised or only-text counterparts, concurrent studies \cite{mlpproj,rethinking,losses_transfer,noreasonnosup} have pointed out that transfer learning from supervised pre-training should be done carefully, as specific optimization criteria and network architectures can substantially impact its performance. Concretely, using softmax cosine similarities, trainable temperature scaling, and an MLP projection during pre-training, as commonly used in contrastive pre-training objectives, are key factors for the proper transferability of such models \cite{noreasonnosup}.

\noindent\textbf{\textit{Large-scale vision models in CXRs}.} Transfer learning from natural to medical domains, and in particular to radiography images, has been a largely adopted and successful strategy \cite{nih} that speeds up convergence and discriminative performance when the training data is limited \cite{transfusion}. To bridge the gap between natural and radiology domains, leveraging large unsupervised datasets via self-supervised learning \cite{bigselfsupmodelscxr,selfsupcxr2} has been exhaustively explored. More recently, the emergence of open-access datasets with radiology reports, i.e., MIMIC \cite{mimic}, has fueled the progress of multimodal models. For example, pre-trained models such as ConVIRT \cite{convirt} and REFERS \cite{refers} demonstrated that incorporating semantic information via language leads to better transferrable features, whereas CheXzero \cite{chexzero} showed radiologist-level performance zero-shot disease recognition. Different strategies are currently being explored to improve pre-training, which include spatial alignment enhancement, i.e., GLoRIA \cite{gloria} and MGCA \cite{mgca}, masking \cite{mrm}, or using soft similarity matrices \cite{sat}. On the other hand, BioViL \cite{biovil} instead focuses on improving text understanding using domain-specific pre-training of the text encoder. Furthermore, a relevant body of recent literature \cite{medclip,cxrclip,ked} explores the integration of supervised labeled datasets to provide larger-scale models. For example, MedCLIP \cite{medclip} proposed aligning unpaired images and texts through labels, via an asymmetrical soft similarity matrix. CXR-CLIP \cite{cxrclip} transforms categorical supervision to text using prompt templates. MedKLIP \cite{medklip} and KED \cite{ked} incorporate domain knowledge and explicitly align the learned representations in the label space. Despite the great efforts devoted to visual-language learning, supervised (i.e., unimodal) pre-training has been surprisingly overlooked, and its potential compared to vision-language models remains unexplored. 

\noindent\textbf{\textit{From text to labels in radiology reports}.} Supervision in chest radiographs naturally comes from text descriptions, which are carried out during clinical routine. These can be accessed in massive amounts from clinical records, and serve as a source to avoid time-consuming image labeling from experienced radiologists. Thanks to the joint effort between radiologists and NLP scientists, several named-entity recognition (NER) tools have been developed, such as Negbio \cite{negbio} or Chexpert-labeler \cite{chexpert}, which are able to extract labels, e.g., diseases and lesions, from text reports. NER algorithms have become the \textit{de facto} solution for labeling large-scale CXR datasets, such as NIH \cite{nih}, CheXpert \cite{chexpert}, MIMIC \cite{mimic}, or PadChest \cite{padchest}. Although these labels could be imperfect, NER algorithms are highly data-efficient \cite{chexpertplus}. Moreover, current entity extraction methods are validated on a wide number of conditions (e.g., 14 for CheXpert \cite{chexpert}, 20 for NIH \cite{nih_lt}, or 96 for PadChest \cite{padchest}). Hence, NER methodologies are continuously improving, and current solutions such as RadGraph \cite{radgraph}, RadText \cite{radtext}, or X-Raydar-NLP \cite{lancet_mlplabels} show promising capabilities. 

\section{Methodology}
\label{sec:methods}

\subsection{Preliminaries}

\begin{figure*}[t!]
\begin{center}
\includegraphics[width=.90\textwidth]{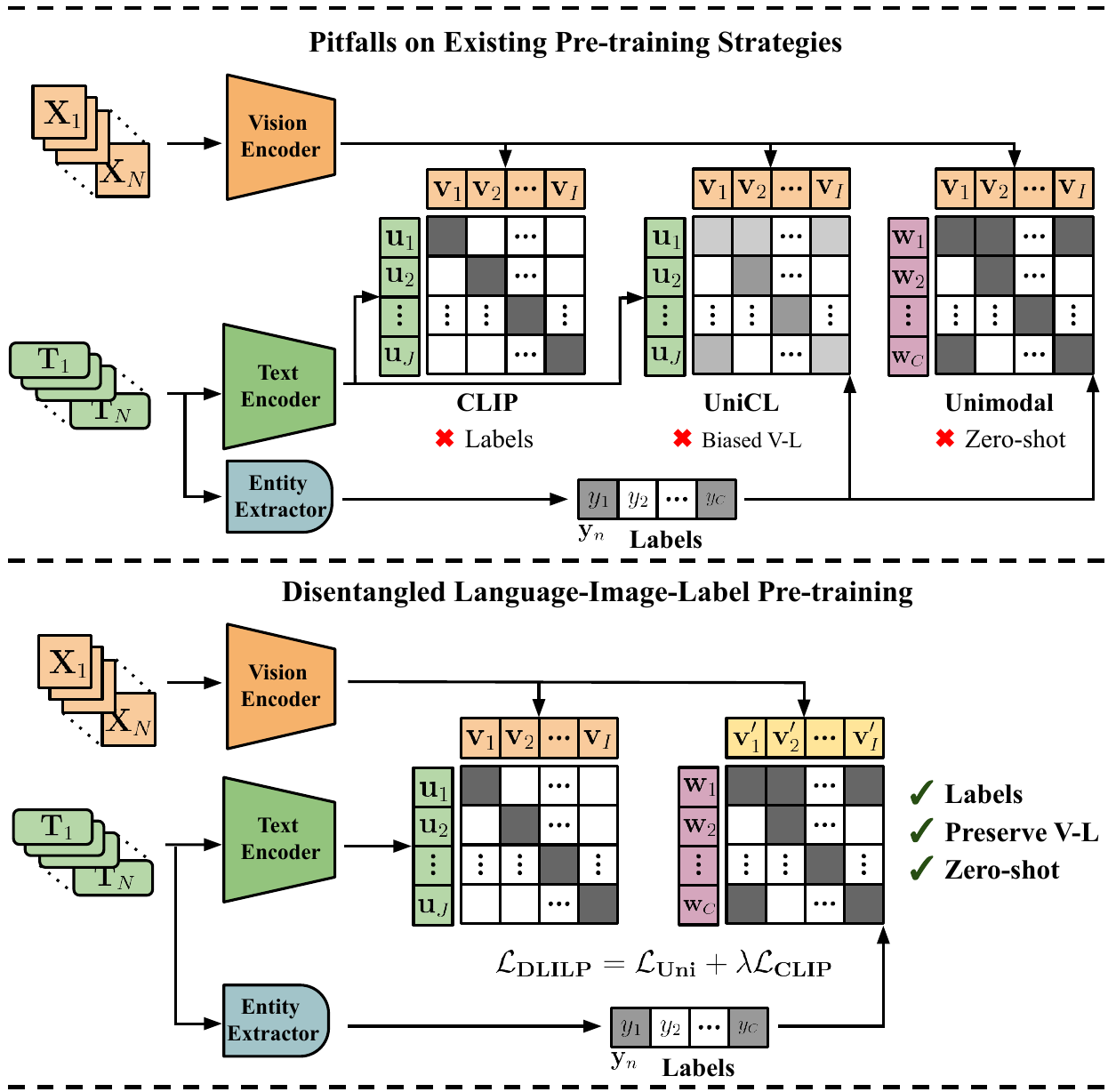}
\caption{\textbf{Training transferable vision models}. Radiology reports include text descriptions, from which labels are extracted through entity extractor methods. Previous methods struggle to align language-image-label information without compromising zero-shot generalization — see Section \ref{sec:pitfalls}. We propose \textbf{DLILP}, a \textbf{D}isentangled \textbf{L}anguage-\textbf{I}mage-\textbf{L}abel \textbf{P}re-training that exploits text and label supervision in separate feature projections, described at Section \ref{sec:dlilp}.}
\label{fig:summary}
\end{center}
\end{figure*}

\paragraph{\textbf{Problem setup.}} We define a quadruplet-wise data format, that generally describes the information available in an assembly of $N$ chest X-ray samples, with text and label supervision, $\mathcal{D}_{ILT}=\{(\rmX_n, \rmY^{\img}_{n}, \rmt_{n}, \rmY^{\textCustome}_{n})\}_{n=1}^{N}$. $\rmX \in \real^{\Omega}$ denotes a CXR 2D image, with $\Omega$ its spatial domain, and $\rmt \in \mathcal{T}$ its associated text description. Furthermore, $\rmY = (y_{1}, ... , y_{c}, ..., y_{C}) $ is a multi-label vector for a set of $C$ \textit{base} categories, such that $y_{c} \in \{0, 1\}$. Note that for one sample $n$, the label information associated with the image, $\rmY^{\img}_{n,c}$, and text description, $\rmY^{\textCustome}_{n,c}$, might be different. $\rmt_{n}$ represents an individual sentence of the whole radiology report. Thus, an individual text description can represent semantic information related only to a subset of the categories that are found in the image. Given an assembly of datasets, $\mathcal{D}$, the objective is to \textbf{learn a strong visual representation model, specialized for CXR image understanding (see Fig. \ref{fig:summary})}.

\paragraph{\textbf{Dual-encoder architectures.}} Let $\theta = \{\theta_f(\cdot), \theta_p(\cdot)\}$ denote the vision encoder, with $\theta_f(\cdot)$ a feature extractor and $\theta_p(\cdot)$ a projection head. The feature extractor $\theta_f(\cdot)$ yields a vision feature representation $\tilde{{\mathbf v}} \in \real^{D_{{\mathbf v}}}: \tilde{{\mathbf v}}_i=\theta_f(\rmX_i)$ of an input image $\rmX_i$, with $D_{{\mathbf v}}$ the dimension of the visual feature space. Similarly, let $\phi = \{\phi_f(\cdot), \phi_p(\cdot)\}$ denote the text encoder, $\phi_f(\cdot)$ being a feature extractor and $\phi_p(\cdot)$ a projection head. The feature extractor $\phi_f(\cdot)$ provides a text embedding $\tilde{{\mathbf u}} \in \real^{D_{{\mathbf u}}}: \tilde{{\mathbf u}}_j=\phi_f(\rmt_j)$ of an input text $\rmt_j$, with $D_{{\mathbf u}}$ denoting the dimensionality of text features. Each of the projection heads, $\theta_p(\cdot)$ and $\phi_p(\cdot)$, maps the independent modality representations into a joint unit hyper-sphere space: ${\mathbf v} = \frac{\theta_p(\tilde{{\mathbf v}})}{||\theta_p(\tilde{{\mathbf v}})||}$ and ${\mathbf u} = \frac{\phi_p(\tilde{{\mathbf u}})}{||\phi_p(\tilde{{\mathbf u}})||}$. In this normalized space, the similarity between image $\rmX_i$ and text description $\rmt_j$ is evaluated by the cosine similarity, ${\mathbf v}^{\top}_i {\mathbf u}_j$, where $\top$ denotes the transpose operator. Optimizing dual-encoder architectures jointly relies on constraining the learned representations to match their textual counterparts and dis-match unpaired ones. The learning process is usually performed in mini-batched stochastic gradient descent. In each step, a batch of indices is randomly retrieved from the assembly dataset, such that $\mathcal{B} \subset \{1, \dots, N\}$.

\subsection{Pitfalls on existing pre-training strategies}
\label{sec:pitfalls}

\paragraph{\textbf{CLIP \cite{clip}.}} Designed for image-text datasets, the learning objective aims to guide paired data to produce similar representations and push-away embedding representations from any unpaired image-text or text-image pair. The one-to-one mapping considers a bidirectional contrastive learning objective, $\mathcal{L}_{\clip} = \mathcal{L}^{\itot}_{\clip} + \mathcal{L}^{\ttoi}_{\clip}$, whose components are defined as:
\begin{align}
\label{eq:clip_i2t}
   & \mathcal{L}^{\itot}_{\clip} (\theta,\phi,\tau | \mathcal{B}) = - \sum_{i \in \mathcal{B}} \text{log} \frac{\text{exp}({\mathbf v}_{i}^{T} {\mathbf u}_{i}/\tau)}{\sum_{j\in\mathcal{B}}\text{exp}({\mathbf v}_{i}^{T} {\mathbf u}_{j}/\tau))} ,  \\
   & \mathcal{L}^{\ttoi}_{\clip} (\theta,\phi,\tau | \mathcal{B}) = - \sum_{j \in \mathcal{B}} \text{log} \frac{\text{exp}({\mathbf v}_{j}^{T} {\mathbf u}_{j}/\tau)}{\sum_{i\in\mathcal{B}}\text{exp}({\mathbf v}_{i}^{T} {\mathbf u}_{j}/\tau))}. 
\end{align}
Even though CLIP loss has proven to be a powerful tool for leveraging large-scale datasets with associated text supervision with minimum supervisory effort, it lacks the fine-grained information that can be found in the form of labels, which the text encoder is assumed to learn. While this does not pose any particular problem in general vision problems, in specialized domains such as medical imaging, with limited data and complex semantics, the text encoder struggles to encode this information efficiently.

\paragraph{\textbf{UniCL \cite{unicl}}} attempts to unify the learning objective across image, text, and label spaces. This is done by modifying the one-to-one similarity matrix in CLIP to a soft-labeled target, by positively pairing images and texts with their labeled categories. The overall training objective, $\mathcal{L}_{\unicl} = \mathcal{L}^{\itot}_{\unicl} + \mathcal{L}^{\ttoi}_{\unicl}$, is defined as:
\begin{align}
   & \mathcal{L}^{\itot}_{\unicl} (\theta,\phi,\tau | \mathcal{B}) = - \sum_{i \in \mathcal{B}} \frac{1}{|P_{\itot}(i)|} \sum_{i' \in P_{\itot}(i)} \text{log} \frac{\text{exp}({\mathbf u}_{i}^{T} {\mathbf v}_{i'}/\tau)}{\sum_{j\in\mathcal{T}_B}\text{exp}({\mathbf u}^{T}_i {\mathbf v}_j/\tau)} , \label{eq:i2t} \\
   & \mathcal{L}^{\ttoi}_{\unicl} (\theta,\phi,\tau | \mathcal{B}) = - \sum_{j \in \mathcal{B}} \frac{1}{|P_{\ttoi}(j)|} \sum_{j' \in P_{\ttoi}(j)} \text{log} \frac{\text{exp}({\mathbf u}_{j'}^{T} {\mathbf v}_{j} / \tau)}{\sum_{i\in\mathcal{X}_B}\text{exp}({\mathbf u}^{T}_i {\mathbf v}_j / \tau)}, \label{eq:t2i}
\end{align}

\noindent where $|\cdot|$ denotes the cardinality of a given set, and $P_{\itot}(i)$ and $P_{\ttoi}(j)$ represent indices of positive-paired cross-modal representations for each image and text in the batch $\mathcal{B}$, respectively. For the multi-label scenario in CXRs, aligned pairs should contain at least one overlapping category, such that:
\[P_{\itot}(i) = \{ i'|(i' \in \mathcal{B}, \exists c | y^{\textCustome}_{i',c} = y^{\img}_{i,c} = 1) \},\]
\[P_{\ttoi}(j) = \{ j'|(j' \in \mathcal{B}, \exists c | y^{\img}_{j',c} = y^{\textCustome}_{j,c} = 1) \}.\]
Although UniCL loss encourages learning both a discriminative and semantic-rich feature space, our empirical evidence (see Section \ref{sec:results_1}, \textit{Observation 2}) suggests that \textbf{label information biases vision-language alignment}. In the case of using a reduced set of labeled categories, as is usually the case in medical domains, the learned representations might fail to capture other information contained in text descriptions, thus worsening their discriminative performance on unseen scenarios during label alignment, i.e., zero-shot predictions.

\paragraph{\textbf{Unimodal supervised learning.}} A classical alternative to pre-train a large-scale vision model using labeled datasets is standard supervised pre-training. In this case, the text encoder is replaced by a linear embedding layer $\tilde{\rmW}^{C\times D_{p}}$, with $D_{p}$ the dimensionality of the projection layer of the visual encoder. In addition, class prototypes are $\ell_{2}$-normalized, such that ${\rmW} = \frac{\tilde{\rmW}}{||\tilde{\rmW}||}$. In the multi-label scenario, class-wise scores are computed using the sigmoid activation function, $\hat{y} = \sigma({\rmW}^{\top} {\mathbf v} / \tau)$, and learning is driven by the binary cross entropy loss:
\begin{align}
\label{eq:unimodalloss}
\mathcal{L}_{\uni}(\theta,\tau, \rmW | \mathcal{B}) = - \sum_{i \in \mathcal{B}} \frac{1}{C}\sum_{c} ( y^{\img}_{i,c} \cdot \text{log}(\hat{y}_{i,c}) + (1-y^{\img}_{i,c}) \cdot \text{log}(1-\hat{y}_{i,c})).
\end{align}
This solution is largely more computationally efficient as it does not involve using a text encoder. In addition, it does not require prompt engineering for generalization on the base categories. A limitation, however, is that only-vision (i.e., unimodal) models lack the capability of zero-shot predictions in novel categories. 

\subsection{Disentangled Language-Image-Label Pre-training}
\label{sec:dlilp}

To address the limitations of label alignment in vision-language pre-training, we propose a \textbf{D}isentangled \textbf{L}anguage-\textbf{I}mage-\textbf{L}abel \textbf{P}re-training (DLILP) strategy.

\noindent\textbf{\textit{Training}.} Image-label and image-text supervision are incorporated into different subspaces of the learned vision representation. In particular, label supervision is driven by the cross-entropy loss, similar to the Unimodal pre-training, whereas we adopt CLIP loss for image-text alignment. To do so, we define two different projection layers, $\theta_p^{\il}$ and $\theta_p^{\it}$, which produce $\ell_{2}$-normalized feature spaces. Formally, the DLILP optimization criteria can be defined as follows:
\begin{align}
\label{eq:dlilp}
\mathcal{L}_{\dlilp} = \mathcal{L}_{\uni}(\{\theta_f, \theta_p^{\il}\},\tau^{\il}, \rmW | \mathcal{B}) + \lambda \cdot \mathcal{L}_{\clip}(\{\theta_f, \theta_p^{\it}\},\phi,\tau^{\it} | \mathcal{B}),
\end{align}

\noindent where $\lambda$ is a blending hyper-parameter that balances the relative importance of vision-language and vision-label pre-training. Note that we train separate temperature scaling parameters, $\tau^{\il}$ and $\tau^{\it}$, for each term.

\noindent\textbf{\textit{Inference}.} DLILP allows robust generalization over known categories using the learned class prototypes, $\rmW$, and the image-label projection. In the case of novel categories, zero-shot predictions using engineered text prompts can also be computed, using the unbiased image-text projection of the vision encoder, and the prototypes obtained using the text encoder.

\section{Experiments}
\label{sec:experiments}

\subsection{Setup}

\noindent\textbf{\textit{Datasets}.} Frontal chest X-ray open-access datasets are employed to train and evaluate the transferability of pre-trained models. Table \ref{datasets_assembly} depicts a summary, and \appensecref{supp:datasets} specific details. For the \textbf{pre-training} stage, we used large datasets such as MIMIC (M) \cite{mimic} and CheXpert (C) \cite{chexpert}. The 14 \textit{base categories} ($\mathcal{B}$) labeled in the CheXpert dataset are considered for label alignment during pre-training. PadChest (P) \cite{padchest}, containing 84 different findings, is used when specified. Labels are extracted from text-only datasets using CheXpert-labeler \cite{chexpert}. For label-only datasets, the text is obtained using a template as in \cite{unicl,chexzero}. To \textbf{evaluate} the capabilities of the resulting models, we used seven different datasets: MIMIC \cite{mimic}, CheXpert \cite{chexpert}, SSIM \cite{ssim}, RSNA \cite{rsna}, NIH \cite{nih}, VinDr \cite{vindr}, and COVID \cite{chexpert}. Some of these datasets include \textit{novel diseases} ($\mathcal{N}$), which have not been explicitly used during image-label alignment in the pre-training.

\begin{table}[t!]
\centering
\caption{\textbf{Frontal Chest X-ray datasets assembly.} We compiled open-access datasets for training and evaluation. Green-colored categories indicate \textcolor{teal}{novel classes} not explicitly used during CheXpert and MIMIC pre-training.}
\label{datasets_assembly}
\begin{tabular}{lrrcl}
\midrule
\multicolumn{1}{l}{\textbf{Pre-train}}       & \multicolumn{1}{c}{\textbf{\#Imgs}} & \multicolumn{1}{c}{\textbf{Text}} & \multicolumn{1}{c}{\textbf{\#C}} & \multicolumn{1}{l}{\textbf{Categories}} \\
\midrule
CheXpert (C)\cite{chexpert}  & 191,026 & \multicolumn{1}{c}{-}                                      & 14 & \multirow{2}{*}{\begin{tabular}[c]{@{}l@{}}[NF, ECard, Card, LLes, LOp, Edem, Cons,  \\ PnMo, Atel, PnTh, PlEff, PlOt, Fract, Dev]  \end{tabular}} \\
MIMIC (M)\cite{mimic}    & 154,595 & \multicolumn{1}{c}{\textbf{\checkmark}} & 14 & \\
PadChest (P)\cite{padchest}    & 96,201 & \multicolumn{1}{c}{-} & 84 & (see \href{https://github.com/jusiro/DLILP/blob/main/cxrvlms/modeling/constants.py}{{code}}) \\
\midrule
\multicolumn{1}{l}{\textbf{Evaluation}}     & \multicolumn{1}{c}{\textbf{\#Train}} & \multicolumn{1}{c}{\textbf{\#Test}} & \multicolumn{1}{c}{\textbf{\#C}} & \multicolumn{1}{l}{\textbf{Categories}} \\
\midrule
CheXpert$_{5\times200}$     & 1,000 & 1,000 & 5 & [Atel, Card, Cons, Edem, PlEff]        \\
MIMIC$_{5\times200}$        & 1,000 & 1,000 & 5 & [Atel, Card, Cons, Edem, PlEff]        \\
RSNA \cite{rsna}            & 8,400 & 3,600 & 2 & [NF, PnMo] \\
SSIM \cite{ssim}         & 4800 & 1200  & 2 & [NF, PnTh] \\
COVID \cite{covid1,covid2}  & 1,200 & 4,000 & 4 & [Normal, \textcolor{teal}{COVID}, N-COVID PnMo, LOp] \\
NIH-LT\cite{nih,nih_lt}  & 920 & 920 & 20 & \multirow{2}{*}{\begin{tabular}[c]{@{}l@{}}[Atel, Card, PlEff, \textcolor{teal}{Inf}, \textcolor{teal}{Mass}, \textcolor{teal}{Nod}, PnMo, \\ PnTh, Cons, Edem, \textcolor{teal}{Emph}, \textcolor{teal}{Fib}, \textcolor{teal}{PlThi}, \\ \textcolor{teal}{PnPer}, \textcolor{teal}{PnMed}, \textcolor{teal}{SubEm}, \textcolor{teal}{TAor}, \textcolor{teal}{CalAor}, NF]  \end{tabular}} \\ 
  &  &  &  & \\ 
  &  &  &  & \\
VinDr \cite{vindr}          & 2,000 & 2,000 & 5 & [NF, \textcolor{teal}{Bro}, \textcolor{teal}{BrPn}, \textcolor{teal}{BrLi}, PnMo]        \\
\bottomrule
\end{tabular}
\end{table}

\noindent\textbf{\textit{Vision-language architecture}.} We designed both encoders following relevant prior literature in the topic \cite{convirt,medclip,medklip,cxrclip}. In particular, we used ResNet-50 \cite{resnet} pre-trained on ImageNet \cite{imagenet} as a vision encoder, $\theta$, and BioClinicalBERT \cite{bioclinicalbert} as the text encoder. All feature projections, i.e., $\theta_p(\cdot)$ and $\phi_p(\cdot)$, $\theta_p^{\il}$ and $\theta_p^{\it}$, are linear layers of $512$ output features, following prior works \cite{clip,medclip}.

\noindent\textbf{\textit{Large-scale training}.} The vision and text encoders are trained using a batch size of $128$ images of $224\times224$ pixels. AdamW is used as the optimizer, with a weight decay of $10^{-5}$, and a base learning rate of $10^{-4}$. Cosine scheduler decay is applied for $30$ epochs, with an initial first warm-up epoch. The $10\%$ of the training subset is sampled for validation. The same data augmentation used in prior related literature \cite{medclip} is applied: random horizontal flips, rotations up to $5$ degrees, scaling between $[0.9, 1.1]$ factor ranges, and color jittering with brightness and contrast ratios from $[0.8, 1.2]$. Validation loss is monitored epoch-wise during training, and early-stopping is applied with a margin of $5$ epochs, saving the best model weights. For DLILP, the $\lambda$ hyper-parameter is set to $0.1$.

\noindent\textbf{\textit{Transferability}.} The transfer capabilities of each pre-training strategy are evaluated in the zero- and few-shot regimes. \textbf{a) Zero-shot}: for CLIP and UniCL frameworks, text-driven class-wise prototypes are obtained using an assembly of text prompts, as in \cite{medclip}. For the Unimodal pre-training, only zero-shot classification on known categories is possible by retrieving the class weights of the target categories, $\mathbf{W}_c$. For DLILP, we follow a hybrid approach, using image-text or image-label projections, depending on whether the target category is known, as detailed in Section \ref{sec:dlilp}. Finally, class-wise scores are obtained in all cases by computing softmax cosine similarity between class prototypes and projected vision features. \textbf{b) Linear probing}: we use the vision features before the projection layer, $\tilde{{\mathbf v}}$, to train a linear classifier. Concretely, the same solver proposed in CLIP \cite{clip} is used. The adaptation is performed in the popular few-shot regime \cite{shakeri2024few}, in which only $K=\{1, 2, 4, 8, 16\}$ images per class are available for adaptation.

\noindent\textbf{\textit{Evaluation protocol}.} Experiments are repeated using $5$ different random seeds. When evaluating using base-only ($\mathcal{B}$) or novel-only ($\mathcal{N}$) target diseases classification, the corresponding subset of categories is separated for adaptation and evaluation. Average class-wise accuracy (ACA) is used as a metric, as in \cite{medclip}.

\noindent\textbf{\textit{Baselines}.} The transferability of the proposed strategies is compared to relevant SoTA models. We gathered the pre-trained weights (when available) and conducted transferability experiments. Concretely, GlorIA \cite{gloria}, MedCLIP \cite{medclip}, BioVIL\cite{biovil}, MedKLIP \cite{medklip}, and KED \cite{ked} are included. MedCLIP, MedKLIP, and KED, include label alignment during model pre-training. In particular, MedCLIP pre-training follows a training objective similar to UniCL's.
 
\subsection{Main results}
\label{sec:results_1}

\begin{figure*}[t!]
    \begin{center}
        \begin{tabular}{cc}

         \includegraphics[width=.45\linewidth]{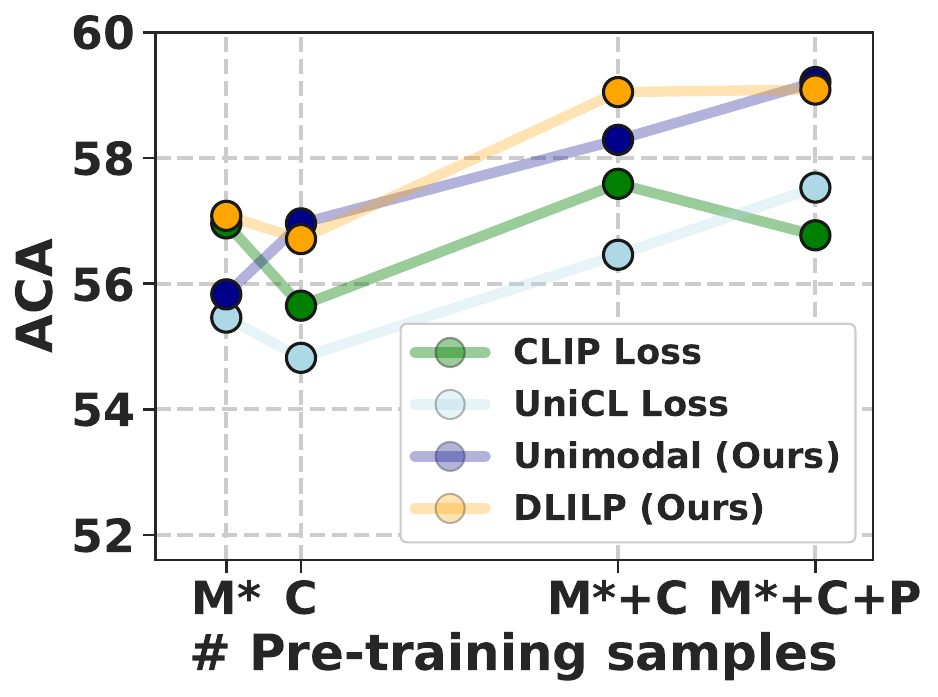} & 
         \includegraphics[width=.45\linewidth]{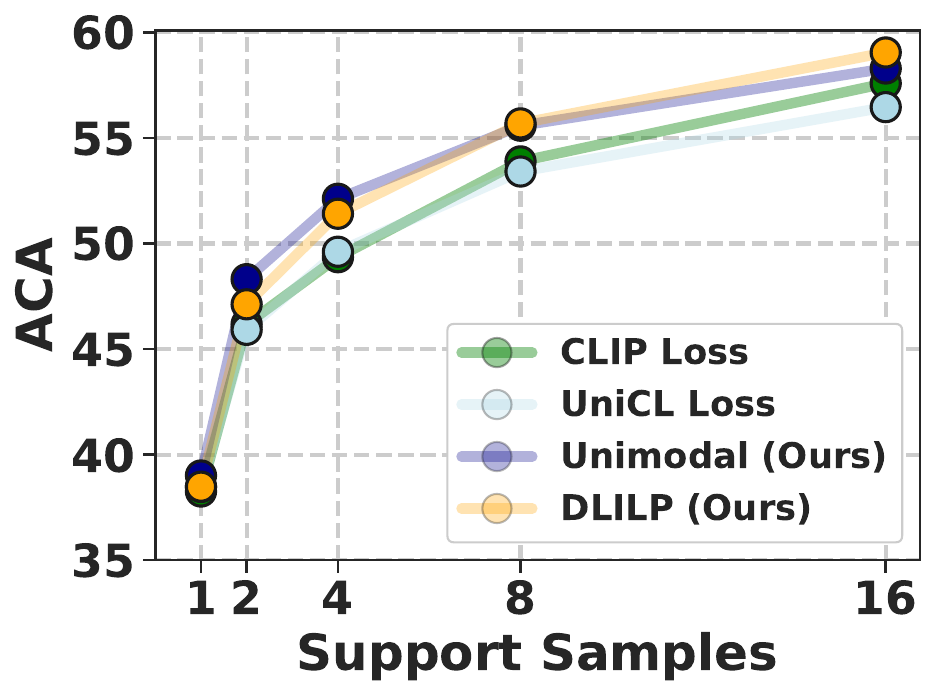} \\
         \textbf{(a) Data scalability} & \textbf{(b) Few-shot adaptation} \\

        \end{tabular}
        \caption{\textbf{Transferability}. (a) Effect of increasing pre-training data (K=16); (b) Few-shot adaptation. Average for 7 tasks. M: MIMIC; C: CheXpert; P: PadChest.}
        \label{fig:transfe}
    \end{center}
\end{figure*}

\noindent\textbf{\textit{Observation 1: Unimodal leads to more scalable transferability than existing vision-language models}.} We compare the few-shot transferability of the different pre-training strategies over the 7 downstream datasets. Fig. \ref{fig:transfe}(a) includes transferability results with increasing pre-training data, which show that CLIP loss struggles to scale properly when adding label-only datasets (see M+C to M+C+P). In contrast, supervised cross-entropy constantly improves w.r.t. the amount of data available (see M+C or M+C+P). Also, Fig. \ref{fig:transfe}(b) shows few-shot transferability results when pre-trained with M+C datasets for different shots. Again, Unimodal offers better adaptation than CLIP and UniCL ($K\geq2$). 

\begin{table}[t!]
\caption{\textbf{Generalization/Transferability results}. Performance of different pre-training strategies disentangling known ($\mathcal{B}$) and new findings ($\mathcal{N}$).}
\label{novel_base}
\centering
\begin{tabular}{lccccccccccc}
        \toprule
        & CheXp & MIMIC & SSIM & RNSA & \multicolumn{2}{c}{NIH$_{LT}$} & \multicolumn{2}{c}{VinDR} & \multicolumn{3}{c}{\textbf{Avg.}} \\ \cmidrule(l){6-7} \cmidrule(l){8-9} \cmidrule(l){10-12}                             & $\mathcal{B}$ & $\mathcal{B}$ & $\mathcal{B}$ & $\mathcal{B}$ & $\mathcal{B}$ & $\mathcal{N}$ & $\mathcal{B}$ & $\mathcal{N}$ & $\mathcal{B}$ & $\mathcal{N}$ & Avg. \\
        \midrule 
        \multicolumn{12}{l}{\textbf{(a) Zero-shot generalization}} \\ \hdashline
        CLIP                           & 51.50 & 49.70 & 77.80 & 63.04 & 40.98 & 29.10  & 68.66 & 32.20 & 58.61          & \textbf{30.65} & 44.63 \\
        UniCL                          & 45.40 & 46.60 & 75.30 & 90.86 & 57.66 & ~~9.10 & 73.16 & 42.20 & 64.83          & 25.65          & 45.24 \\
        \rowcolor{Gray}Unimodal        & 42.80 & 47.40 & 77.20 & 94.60 & 61.70 &     -  & 65.80 &     - & \textbf{64.92} &     -          &     - \\
        \rowcolor{Gray}DLILP           & 49.50 & 48.60 & 77.90 & 93.50 & 60.80 & 29.10  & 54.20 & 31.10 & 64.08          & 30.10          & \textbf{47.09} \\
        \midrule 
        \multicolumn{12}{l}{\textbf{(b) Linear probing transferability ($K=16$)}} \\ \hdashline
        CLIP                           & 54.50 & 49.60 & 69.10 & 93.20 & 46.52 & 32.50  & 71.68 & 38.20 & 64.10          & 35.35           & 49.73  \\
        UniCL                          & 53.10 & 50.90 & 65.58 & 93.78 & 46.50 & 27.52  & 71.32 & 37.54 & 63.53          & 32.53           & 48.03  \\
        \rowcolor{Gray}Unimodal        & 54.20 & 53.70 & 67.68 & 94.36 & 47.16 & 33.20  & 75.34 & 37.44 & 65.41          & 35.32           & 50.37  \\
        \rowcolor{Gray}DLILP           & 55.60 & 54.50 & 72.74 & 93.82 & 50.66 & 32.24  & 71.36 & 40.76 & \textbf{66.45} & \textbf{36.50}  & \textbf{51.48}  \\
        \bottomrule
\end{tabular}
\end{table}

\noindent\textbf{\textit{Observation 2: Label alignment during vision-language pre-training might produce biased joint representations}.} We now study the capability of each pre-training strategy to generalize to novel categories. Results in Table \ref{novel_base} disentangle the zero-shot and linear probing performance between base and new findings. \textbf{a) Zero-shot}: UniCL archives average improvements ($+6.2\%$) compared to the original CLIP on known categories thanks to the label information incorporated. However, it largely degrades the performance when evaluated on novel categories ($-5.0\%$). Interestingly, Unimodal pre-training offers the best results for base categories. Note that this strategy is more computationally efficient since does not require training any text encoder. Also, this method does not require heuristic prompt engineering to define the zero-shot text prompts properly, thanks to using learned prototypes. \textbf{b) Linear probing}: Again, Unimodal pre-training is largely a competitive solution, with slightly better overall performance compared to CLIP loss ($+0.6\%$). Note that UniCL loss does not show any benefits, reinforcing the inconvenience of this label-driven loss.

\noindent\textbf{\textit{Observation 3: The zero-shot capabilities of CXR vision-language models have been overestimated}.} Prior literature, i.e., MedCLIP \cite{medclip} and MedKLIP \cite{medklip}, have defended the effectiveness of vision-language pre-training to generalize to unseen diseases thanks to text-driven predictions.
These experiments have been typically carried out in the COVID dataset by differentiating between normal and COVID scans using text descriptions (see Table \ref{tab:covid}). However, this description (see * in Table \ref{tab:covid}) contains lesions that appeared in the pre-training stage. These findings (i.e., opacities and consolidations) are nonspecific \cite{covidfindings} and may be correlated with other lung conditions. We extend this benchmark to four categories available within the same dataset (see Table \ref{datasets_assembly}) in Table \ref{tab:covid}. In this scenario, the overall performance degrades greatly\footnote{Note that lung opacities might present overlap with pneumonia labels. Hence, we also provide results for only 3-classes in \appensecref{supp:results}, with similar conclusions.}. More interestingly, following the same zero-shot prediction strategy, we can obtain class-wise prototypes for the Unimodal pre-training by selecting the weights corresponding to the findings in the description. The visual prompt for COVID would be the average embedding between the pre-trained prototypes for opacity and consolidation. Surprisingly, this option outperforms the designed text prompts in VLMs. These observations, combined with the limited generalization observed for novel diseases in Table \ref{novel_base}(a), question the advancements claimed in recent literature for open-vocabulary generalization.

\begin{table}[t!]
\centering
\caption{\textbf{Zero-shot on COVID dataset}.} \label{tab:covid}
    \begin{tabular}{llcccccc}
    \toprule
     & & MedCLIP &  MedKLIP & CLIP & UniCL & Unimodal & DLILP \\ \midrule
     \multirow{2}{*}{2-class} & Disease name         &  74.1 &  51.8 & 69.6 & 80.5 & - & 77.0   \\
     & Description*                           &  78.8 &  82.9 & 74.2 & 83.7 & \textbf{85.1} & 81.6    \\ \midrule
     \multirow{2}{*}{4-class} & Disease name         &  40.5 &  20.2 & 32.7 & 45.5 & -  & 36.6   \\
     & Description*                           &  42.9 &  32.5 & 48.8 & 44.8 & \textbf{51.6} & 50.0   \\
    \bottomrule 
    \multicolumn{8}{c}{{{*"\textit{patchy or confluent, band like ground-glass \textbf{opacity} or \textbf{consolidation}"}}}} \\
    \end{tabular}
\end{table}

\noindent\textbf{\textit{DLILP performance}.} Although existing vision-language pre-training alternatives offer limited contributions compared to Unimodal, the proposed DLILP objective shows interesting properties. First, DLILP shows better scalability concerning data integration over baseline VLMs (see Fig. \ref{fig:transfe}(a)). Second, DLILP demonstrates robust zero-shot generalization across both base and new categories (see Table \ref{novel_base}), with the best average performance across both sets for both zero-shot ($+1.9\%$ over UniCL) and few-shot ($+3.5\%$ over UniCL).

\noindent\textbf{\textit{SoTA comparison}.} Table \ref{sota_comparison} is introduced without base/new disentanglement since prior models might present different base categories (e.g., MedKLIP \cite{medklip} or KED \cite{ked}). Unimodal obtains the best results, whose average improvements w.r.t. top competitors range $[2.6\%, 5.6\%]$. This observation applies also to models including label information, such as MedCLIP \cite{medclip}, MedKLIP \cite{medklip}, or KED \cite{ked}.
 
\begin{table}[t!]
\centering
\caption{\textbf{Available vision-language models transferability.} Linear probing results ($K=16$) for SoTA pre-trained models.}
\label{sota_comparison}
\begin{tabular}{llcccccccc}
\toprule
Method & Data    & CheXp & MIMIC & SSIM  & RNSA  & NIH   & VinDR & COVID &\textbf{Avg.}  \\
\midrule
MedKLIP \cite{medklip}     & M     & 34.30 & 32.60 & 64.82 & 88.18 & 14.04 & 26.34 & 68.04 & 46.90 \\
KED \cite{ked}         & M     & 42.50 & 40.20 & 66.04 & 92.12 & 19.40 & 26.18 & 73.24 & 51.38 \\
BioVIL \cite{biovil}       & M & 46.70 & 43.80 & 73.68 & 94.08 & 21.22 & 26.20 & 62.46 & 52.59 \\
\rowcolor{Gray}Unimodal                      & M     & 51.80 & 51.30 & 68.04 & 93.42 & 21.20 & 27.68 & 77.40 & 55.83 \\
\rowcolor{Gray}DLILP                         & M     & 53.30 & 52.90 & 69.80 & 93.78 & 25.34 & 26.84 & 77.62 & \textbf{57.08} \\
\midrule
GlorIA \cite{gloria}       & C     & 46.00 & 41.60 & 66.30 & 91.16 & 18.78 & 23.02 & 72.92 & 51.40  \\
\rowcolor{Gray}Unimodal                      & C     & 52.30 & 48.20 & 71.52 & 93.88 & 24.20 & 29.14 & 79.48 & \textbf{56.96} \\
\midrule
MedCLIP \cite{medclip}    & M+C   & 54.40 & 50.50 & 69.48 & 94.20 & 20.98 & 27.80 & 72.30 & 55.67 \\
CXR-CLIP \cite{cxrclip}  & M+C   & 52.20 & 46.10 & 69.34 & 92.00 & 25.90 & 26.26 & 76.82 & 55.52 \\
\rowcolor{Gray}Unimodal                      & M+C   & 54.20 & 53.70 & 67.68 & 94.36 & 26.20 & 30.26 & 81.62 & 58.29 \\
\rowcolor{Gray}DLILP                         & M+C   & 55.60 & 54.50 & 72.74 & 93.82 & 26.72 & 28.98 & 81.02 & \textbf{59.05}  \\
\bottomrule
\end{tabular}
\end{table}

\subsection{Ablation studies}
\label{sec:ablation}

\noindent\textbf{\textit{What features to transfer}?} We evaluate two possibilities: using the features extracted by the vision encoder, $\tilde{v}$, or the ones projected, $v$. Using the first ones improves base CLIP loss transferability ($+2.3\%$), but especially label-driven learning losses, i.e. UniCL ($+2.6\%$), Unimodal ($+2.6\%$), and DLILP ($+3.2\%$).

\noindent\textbf{\textit{DLILP configuration}.} Table \ref{dlilp_config}(a) motivates disentangling image-label and image-text supervisory signals in different projections. 

\noindent\textbf{\textit{On the effect of $\lambda$}.} Table \ref{dlilp_config}(b) studies $\lambda$ in Eq. \ref{eq:dlilp}. Small values of $\lambda$ offer the best base/novel average performance. Comparing these results to Table \ref{novel_base}(b), one could find that $\lambda$ values between 0.1 and 10 offer average gains to all baselines.

\begin{table}[t!]
\centering
\caption{\textbf{DLILP configuration.} Linear probe ($K=16$), across datasets.}
\label{dlilp_config}
\begin{tabular}{ cccc }
            \begin{tabular}{lcc}
            \toprule
            \multicolumn{1}{c}{(a) Projections} & $\{\theta_p\}$  & $\{\theta_p^{\il},\theta_p^{\it}\}$ \\ \midrule
            \textit{Base}                 & 65.2    & \textbf{66.5}$_{(+1.3)}$\textcolor{blue}{$\uparrow$}  \\
            \textit{Novel}                & 35.4    & \textbf{36.5}$_{(+1.1)}$\textcolor{blue}{$\uparrow$}  \\
            Avg.                          & 50.3    & \textbf{51.5}$_{(+1.2)}$\textcolor{blue}{$\uparrow$}  \\
            \bottomrule
            \end{tabular} & & & 
            \begin{tabular}{lcccc}
                \toprule
                \multicolumn{1}{c}{(b) Effect of $\lambda$} & 0    & 0.1 & 1  & 10 \\ \midrule
                \textit{Base}                 & 65.4 & 66.5          & 65.9 & 64.8 \\
                \textit{Novel}                & 35.3 & 36.5          & 36.8 & 36.1  \\
                Avg.                          & 50.4 & \textbf{51.5} & 51.4 & 50.4  \\
                \bottomrule
            \end{tabular} \\
\end{tabular}
\end{table}

\section{Discussion}
\label{sec:conclusion}

This work addresses large-scale pre-training for CXR image classification. In this topic, fine-grained labels extracted with specialized entity extraction methods are usually the only available information. However, current literature mostly focuses on (noisy) vision-language pre-training, following CLIP's popularity. As we observe in this work, current experimental designs mask the actual transferability of such networks, especially w.r.t. novel diseases. Indeed, when properly compared with classical unimodal pre-training, such approaches showcase limited advantages. We would want to emphasize that this work does not aim to neglect the unarguable progress made in multimodal learning, e.g., in related topics such as medical report generation. On the contrary, this paper aims to point out better evaluation designs (e.g. differentiating $\mathcal{B}$/$\mathcal{N}$ conditions) and establish adequate baselines to measure the progress of pre-training strategies in the field, where Unimodal and DLILP are to be taken into consideration.

\begin{credits}
\subsubsection{\ackname} This work was funded by the Natural Sciences and Engineering Research Council of Canada (NSERC). We also thank Calcul Québec and Compute Canada.
\subsubsection{\discintname} The authors have no competing interests to declare that are relevant to the content of this article.
\end{credits}

\bibliographystyle{splncs04}
\bibliography{refs}

\clearpage
\appendix
\setcounter{section}{0}
\renewcommand{\thesection}{\Alph{section}}

\section*{Supplementary Material}

\section{Methodological Details}
\label{supp:methods}

\noindent\textbf{\textit{Addressing partially labeled datasets}.} When assembling different data sources, those might potentially be partially labeled. This means a subset of classes from the total unique $C$ labeled categories might not be labeled for one dataset. To address such a setting, binary cross-entropy is backpropagated uniquely from the labeled categories for each sample via masking. This strategy has been typically followed for training foundation models in medical volumetric segmentation \cite{multitalent,FSEFT}. Formally, let us denote a sample-level vector of annotations, $\rma_c \in \{0, 1\}$, that for each class, its corresponding value is positive if such label is annotated in its source dataset. The Unimodal supervised masked cross-entropy loss is:
\begin{align}
\label{eq:unimodalloss_partial}
\mathcal{L}^{\text{\tiny{part}}}_{\uni}(\theta,\tau, \rmW | \mathcal{B}) = - \sum_{i \in \mathcal{B}} \sum_{c} \frac{\rma_{i,c}}{|\rma_i|} \cdot (( y^{\img}_{i,c} \cdot \text{log}(\hat{y}_{i,c}) + (1-y^{\img}_{i,c}) \cdot \text{log}(1-\hat{y}_{i,c}))),
\end{align}
where $|\cdot|$ indicates cardinality, i.e., the number of labeled classes for each concrete sample. In the particular scenario of using MIMIC, CheXpert, and PadChest during pre-training, we used the partial binary cross-entropy loss because PadChest introduces categories not labeled in the first datasets (see Table \ref{datasets_assembly} and Section \ref{supp:datasets} for details). 

\section{Datasets Details}
\label{supp:datasets}

\noindent\textbf{\textit{Datasets}.} In the following, we provide specific details on the data preparation and partitioning performed to assemble different frontal chest X-ray scanner datasets. A summary is introduced in Table \ref{datasets_assembly}.

\begin{itemize}
    \item[$\circ$] CheXpert \cite{chexpert} is a large dataset that contains 224,316 frontal and lateral chest radiographs of 65,240 patients. This dataset does not provide the original text reports but $14$ labels of relevant clinical conditions that were extracted using refined entity extraction methods. We used the train partition during pre-training and followed \cite{medclip,gloria} for evaluation. Concretely, a multi-class subset so-called CheXpert$_{5\times200}$ is sampled for testing purposes. Concretely, for CheXpert$_{5\times200}$, in particular, the same samples as in \cite{gloria}. This partition includes 200 samples from 5 categories: Atelectasis, Cadiomelagy, Consolidation, Edema, and Pleural Effusion.
    \item[$\circ$] MIMIC \cite{mimic} is a large-scale dataset that includes 257,345 frontal and lateral views with free-text radiology reports. We processed text reports (“Findings” and “Impression” sections) to extract the same 14 labels provided in CheXpert from each radiology report, as previously done in MedCLIP \cite{medclip}. Concretely, we first divided descriptions into individual sentences of at least 10 characters and then used Chexpert-labeler \cite{chexpert} to leverage the entities. We treated uncertain outputs as negatives\footnote{Although other strategies are possible, we select 0s assignment for its simplicity and good performance in \cite{chexpert}. Even though more complex methodologies could be explored, these fall out of the scope of this work.}. Image-level labels were obtained by combining the labels from individual sentences. A "No finding" label is assigned when any entity is detected in the whole text report. Otherwise, findings encountered in individual sentences are assigned to the sample global image-level label. Analogously to CheXpert, we aligned with relevant prior literature \cite{medclip,gloria}, and sampled a MIMIC$_{5\times200}$ subset for evaluation.
    \item[$\circ$] RSNA \cite{rsna} is a collection of frontal chest x-rays with potential pneumonia and non-pneumonia (normal) cases. The challenge focuses on lung opacities detection and grounding. By leveraging the detailed patient information, we sampled a balanced dataset for training (n=8,400) and testing (n=3,600) in the adaptation stage.
    \item[$\circ$] SIIM \cite{ssim} is a dataset to asses the localization of pneumothorax signs in frontal chest x-rays. We leveraged image-level labels and sampled a balanced dataset for training (n=4800) and testing (n=1200).
    \item[$\circ$] COVID-19 \cite{covid1,covid2} is a dataset used in \cite{medclip,medklip} to evaluate the zero-shot capabilities of vision-language models to discriminate novel categories based on text descriptions. This dataset consists of an assembly of different sources, consisting of four categories: normal, COVID, non-COVID viral pneumonia, and lung opacities. In contrast to previous works, which only focus on normal vs. COVID discrimination, we sampled balanced training (n=1,200) and testing (n=4000) subsets, which include all conditions.
    \item[$\circ$] NIH-LT \cite{nih_lt} is a partition of NIH \cite{nih} (a.k.a. ChestX-ray14) dataset with 5 additional labels extracted via an entity extraction algorithm (i.e., Radtex \cite{radtext}), which sum-ups 20 different diseases, from which 11 are unknown during pre-training. We employed this partition to evaluate the capabilities of pre-trained models to face novel conditions. We combined validation and test partitions to leverage a balanced dataset for evaluation (n=920). We omitted the training subset to ensure a balanced transferability dataset since the original NIH-LT is tailored to long-tailed training.
    \item[$\circ$] VinDr-PCXR \cite{vindr} is a dataset containing frontal radiographs from pediatric patients, which suppose a significant domain shift compared to MIMIC and CheXpert, with up to 22 local lesions and 6 diseases labeled by expert radiologists. We combined train and test splits, and following \cite{cxrclip}, we discarded cases labeled "other disease.". Due to the significant class imbalance, we discarded the cases belonging to categories with less than 400 examples. Finally, we gathered a balanced multi-class dataset (n=2,000) with the resultant categories (n=5), which include two base and three novel categories: No Findings,  Pneumonia, Bronchitis, Brocho-pneumonia, and Bronchiolitis.
\end{itemize}

\noindent\textbf{\textit{Categories}.} In the following, we provide the categories and corresponding abbreviations used for training and adaptation of the chest x-rays (CXR) pre-trained models, used in Table \ref{datasets_assembly}. For further details on the categories existing in PadChest datasets, we refer the reader to its original publication in \cite{padchest}. \textbf{Abbreviations:} No Finding (NF), Enlarged Cardiomediastinum (ECard), Cardiomegaly (Card), Lung Lesion (LLes), Lung Opacity (LOp), Edema (Edem), Consolidation (Cons), Pneumonia (PnMo), Atelectasis (Atel), Pneumothorax (PnTh), Pleural Effusion (PleEff), Pleural Other (PlOt), Fracture (Fract), Support Devices (Dev), Normal, COVID, Infiltration (Inf), Mass, Nodule (Nod), Emphysema (Emph), Fibrosis (Fib), Pleural Ticketning (PlThi), Pneumoperitoneum  (PnPer), Pneumomediastinum (PnMed), Subcutaneous Emphysema (SubEm), Tortuous Aorta (TAor), Calcification of the Aorta (CalAor), Bronchitis (Bro), Brocho-pneumonia (BrPn), Bronchiolitis (BrLi). 

\noindent\textbf{\textit{Paired images and text descriptions with different labels}.} As stated in the main manuscript (see Section \ref{sec:methods}), the labels associated to an image, $\rmY^{\img}_{n,c}$, and a paired text description, $\rmY^{\textCustome}_{n,c}$, might differ. This is a particular characteristic of radiology reports. To address the large extent of radiology reports, those are processed using individual sentences, as in MedCLIP \cite{medclip}. Thus, each sentence might reference individual findings. This motivates our experimental setting, which extracts labels from entity extraction methods in MIMIC sentence-wise, as previously detailed. We provide examples of such cases in Table \ref{examples_labels}.

\begin{table}[t!]
\setlength{\tabcolsep}{2pt}
\centering
\caption{\textbf{Examples of image-text-labels triplets.} Image-level labels might not correspond to individual-sentence labels.}
\label{examples_labels}
\scriptsize
\begin{tabular}{lll}
    \toprule
     \textbf{Sentences} &
      \textbf{Text Labels} &
      \textbf{Image Labels} \\ \midrule
\begin{tabular}[c]{@{}l@{}}1. Hazy widespread opacity which could be compatible \\ with a coinciding pneumonia.\\ 2. Pulmonary nodules in the left upper lobe are also\\  not completely characterized on this study.\end{tabular} &
  \begin{tabular}[c]{@{}l@{}} 1. {[}Lung Opacity{]}\\  \\ 2. {[}Lung Lesion{]} \\ \\ \end{tabular} &
  \begin{tabular}[c]{@{}l@{}}{[}Lung Opacity, \\ Lung Lesion{]}\end{tabular} \\ \midrule
\begin{tabular}[c]{@{}l@{}}1. With exception of mild bibasilar atelectasis, the lungs\\  are normally expanded without focal opacity to suggest \\ pneumonia.\\ 2. Heart size is mildly enlarged.\\ 3. There is no pleural effusion or pneumothorax\end{tabular} &
  \begin{tabular}[c]{@{}l@{}}1. {[}Atelectasis{]}\\ \\ \\ 2. {[}Cardiomelagy{]}\\ 3. {[}No Findings{]}\end{tabular} &
  \begin{tabular}[c]{@{}l@{}}{[}Atelectasis,\\  Cardiomelagy{]}\end{tabular} \\
    \bottomrule
    \end{tabular}
\end{table}

\section{Additional Experimental Details}
\label{supp:results}

\noindent\textbf{\textit{Alternative pre-training baselines}.} Currently, the most popular pre-training strategy for chest X-ray scans are vision-language models \cite{convirt,medclip,refers,gloria,medklip,ked,biovil,cxrclip}. Nevertheless, authors in \cite{sup_pret_cxr} explored Unimodal pre-training, using radiology datasets. However, the objective of that paper is to compare this strategy with transfer learning with respect to pre-trained models in natural images, i.e., ImageNet. In addition, authors in \cite{sup_pret_cxr} use supervised contrastive learning, creating multi-view and multi-class positive and negative anchors. It is worth mentioning that such a method has two major limitations in our setting: \textit{i}) SupCons does not use classification head, and hence does not allow zero-shot generalization on known categories; and \textit{ii}) SupCons is not straightforwardly applicable to multi-label data, which is characteristic in CXR datasets, where each sample should be aligned with multiple categories. Note that authors in \cite{sup_pret_cxr} do not provide details on how (\textit{ii}) is addressed. Moreover, such work's Appendix specifies particular implementation details (e.g., using an additional classification head focused on pneumothorax prediction and early stopping based on such head performance) that hinder direct comparisons. Finally, it is worth mentioning that recent studies on supervised pre-training have pointed out that using class-wise prototypes instead of such contrastive loss consistently provides better transferability \cite{noreasonnosup}.

\begin{figure*}[t!]
\begin{center}
\includegraphics[width=0.75\textwidth]{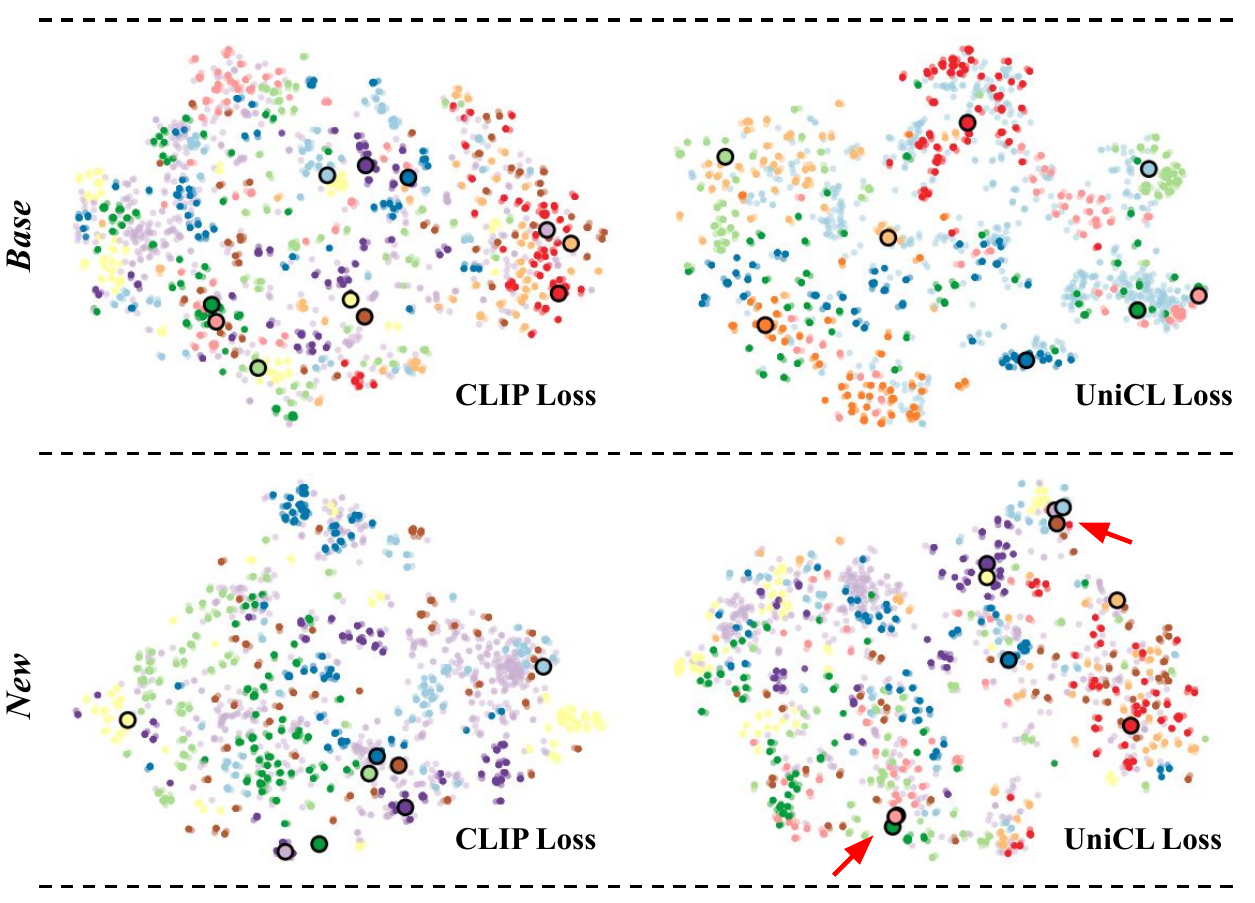}
\caption{\textbf{Pitfalls of UniCL on novel categories}. T-SNE of the embeddings produced after UniCL pre-training on the NIH-LT testing dataset. Large dots represented text prototypes. and small dots represent samples. Each color represents a category. The t-SNE representation shows that UniCL properly aligns labeled categories (\textit{top, right}), but collapses on novel categories \textit{bottom, right}.}\label{qualitative_tsne}
\label{fig:tnse}
\end{center}
\end{figure*}

\noindent\textbf{\textit{Label consistency between image and text prototypes}.} We introduce qualitative examples of the effect of label alignment using UniCL loss. We do so by depicting the t-SNE representation of the visual embeddings obtained at NIH-LT testing data, in Fig. \ref{qualitative_tsne}. Results show that, although the text prototypes are better separated in base categories, UniCL does not show any benefit compared to CLIP loss for unseen findings. This qualitative assessment supports the quantitative results presented in Table \ref{novel_base}(a).

\noindent\textbf{\textit{Extended results on COVID dataset}.} The results depicted in the main paper in Table \ref{tab:covid} tackle the 4-class classification problem. As showcased in Table \ref{datasets_assembly}, the categories tackled are: normal, COVID pneumonia, non-COVID viral pneumonia, and lung opacity. However, the last finding might appear in the general pneumonia cases, which risks overlapping with the targeted categories. In the following, we provide zero-shot performance in Table \ref{tab:covid_extended} only for the first three categories. Results are consistent with our previous findings. Concretely, using hard-crafted prompts for zero-shot generalization to novel diseases in vision-language models does not show any benefit compared to the proposed unimodal prompting strategy based on local findings.

\begin{table}[t!]
\setlength{\tabcolsep}{4pt}
\centering
\caption{\textbf{Zero-shot on COVID dataset - extended results}.} \label{tab:covid_extended}
    \begin{tabular}{llcccc}
    \toprule
     &                                               & CLIP & UniCL & Unimodal      & DLILP \\ \midrule
     \multirow{2}{*}{3-class} & Disease name         & 34.3 & 61.4  & -             & 44.9  \\
     & Description                                  & 45.7 & 53.8  & 55.5          & 51.6  \\
    \bottomrule 
    \end{tabular}
\end{table}

\noindent\textbf{\textit{Extended studies on RNSA}.} The proposed dataset pre-processing for RSNA \cite{rsna} differs from the one employed in \cite{gloria}. Concretely, the authors from GlorIA employed global labels inferred from the presence or absence of local findings. However, the absence of local findings does not necessarily imply that the patient presents a normal scan, since other conditions might be present. Hence, we leveraged the detailed global patient information to create the labels instead. The results obtained for both strategies are depicted in Table \ref{tab:rsna_extended}. These show relative comparisons between pre-training methods consistent in both partitions, but showcase better overall performance if using the detailed patient information.

\begin{table}[h!]
\setlength{\tabcolsep}{4pt}
\centering
\caption{\textbf{Detailed results in RSNA dataset.} Comparison of our proposed partition and the one employed in \cite{gloria}. Zero-shot (ZS) and linear probing (LP) results, the latter using 16 shots.} \label{tab:rsna_extended}
        \begin{tabular}{lcccc}
        \toprule
        \multicolumn{1}{l}{Pre-training} & \multicolumn{2}{c}{Local \cite{gloria}} & \multicolumn{2}{c}{Ours} \\
        \multicolumn{1}{c}{}            & ZS          & LP         & ZS          & LP         \\
        \midrule
        CLIP Loss \cite{clip}     & 56.4 & 77.5  & 63.0 & 93.2 \\
        UniCL Loss \cite{unicl}   & 72.3 & 76.0  & 90.9 & 93.8 \\
        \rowcolor{Gray}Unimodal   & 76.2 & 77.1  & 94.6 & 94.4 \\
        \rowcolor{Gray}DLILP      & 77.3 & 77.3  & 93.5 & 93.8 \\
        \bottomrule 
        \end{tabular}
\end{table}

\noindent\textbf{\textit{Detailed numerical results}.} We introduce the concrete numerical results obtained during the few-shot adaptation of the different explored pre-training strategies on the downstream tasks. Concretely, Table \ref{data_scaling_numerical} introduces numerical results for the 16-shot transferability of different pre-trained models using an increasing number of datasets for training. Table \ref{few_sshot_numerical} depicts figures of merit for an increasing number of shots during adaptation.

\begin{table}[h!]
\setlength{\tabcolsep}{2pt}
\centering
\caption{\textbf{Detailed scalability results.} Linear probing results with $K=16$ shots for the different pre-training strategies with an increasing number of datasets. These results complement visualizations provided in Fig. \ref{fig:transfe}(a).}
\label{data_scaling_numerical}
\scriptsize
\begin{tabular}{llcccccccc}
\toprule
Method & Data    & CheXp & MIMIC & SSIM  & RNSA  & NIH   & VinDR & COVID &\textbf{Avg.}  \\
\midrule
CLIP                                         & M*      & 51.40 & 48.00 & 68.40 & 93.62 & 27.64 & 29.70 & 79.96 & 56.96 \\
UniCL                                        & M*      & 51.20 & 51.10 & 69.30 & 93.74 & 20.26 & 28.26 & 74.38 & 55.46 \\
\rowcolor{Gray}Unimodal                      & M*      & 51.80 & 51.30 & 68.04 & 93.42 & 21.20 & 27.68 & 77.40 & 55.83 \\
\rowcolor{Gray}DLILP                         & M*      & 53.30 & 52.90 & 69.80 & 93.78 & 25.34 & 26.84 & 77.62 & \textbf{57.08} \\
\midrule
CLIP                                         & C       & 50.80 & 47.10 & 70.98 & 93.42 & 22.04 & 27.18 & 78.00 & 55.65 \\
UniCL                                        & C       & 50.50 & 45.70 & 68.78 & 93.00 & 20.70 & 28.54 & 76.54 & 54.82 \\
\rowcolor{Gray}Unimodal                      & C       & 52.30 & 48.20 & 71.52 & 93.88 & 24.20 & 29.14 & 79.48 & \textbf{56.96} \\
\rowcolor{Gray}DLILP                         & C       & 51.60 & 48.10 & 73.02 & 94.58 & 23.38 & 28.40 & 77.92 & 56.71 \\
\midrule
CLIP                                         & M*+C    & 54.50 & 49.60 & 69.10 & 93.20 & 25.76 & 28.34 & 82.66 & 57.59 \\
UniCL                                        & M*+C    & 53.10 & 50.90 & 65.58 & 93.78 & 24.56 & 26.94 & 80.38 & 56.46 \\
\rowcolor{Gray}Unimodal                      & M*+C    & 54.20 & 53.70 & 67.68 & 94.36 & 26.20 & 30.26 & 81.62 & 58.29 \\
\rowcolor{Gray}DLILP                         & M*+C    & 55.60 & 54.50 & 72.74 & 93.82 & 26.72 & 28.98 & 81.02 & \textbf{59.05}  \\
\midrule
CLIP                                         & M*+C+P   & 51.70 & 50.00 & 70.42 & 93.64 & 24.64 & 30.56 & 76.46 & 56.77 \\
UniCL                                        & M*+C+P   & 51.50 & 52.70 & 66.32 & 93.86 & 26.90 & 30.80 & 80.64 & 57.53 \\
\rowcolor{Gray}Unimodal                      & M*+C+P   & 56.00 & 55.20 & 73.84 & 94.00 & 26.12 & 28.48 & 80.86 & \textbf{59.21} \\
\rowcolor{Gray}DLILP                         & M*+C+P   & 51.60 & 53.50 & 68.62 & 93.9  & 30.30 & 28.96 & 79.58 & 58.07 \\
\bottomrule
\multicolumn{10}{l}{M: MIMIC; C: CheXpert; P: PadChest. Image-text datasets indicated by *.}
\end{tabular}
\end{table}

\begin{table}[h!]
\setlength{\tabcolsep}{2pt}
\centering
\caption{\textbf{Detailed few-shot linear probing results.} Transferability results for the different pre-training strategies with an increasing number of shots for adaptation. These results complement visualizations provided in Fig. \ref{fig:transfe}(b). Results using MIMIC and CheXpert datasets for pre-training.}
\label{few_sshot_numerical}
\scriptsize
\begin{tabular}{llcccccccc}
\toprule
Method & Shots    & CheXp & MIMIC & SSIM  & RNSA  & NIH   & VinDR & COVID &\textbf{Avg.}  \\
\midrule
CLIP                                         & \multirow{3}{*}{1-shot}  & 28.50 & 27.80 & 55.20 & 74.5  & 10.52 & 22.34 & 48.88 & 38.25 \\
UniCL                                        &                          & 27.40 & 32.40 & 54.04 & 71.66 & 10.46 & 20.26 & 53.04 & 38.47 \\
\rowcolor{Gray}Unimodal                      &                          & 26.90 & 31.30 & 60.84 & 69.48 & 11.22 & 22.40 & 50.96 & \textbf{39.01} \\
\rowcolor{Gray}DLILP                         &                          & 27.30 & 31.30 & 56.22 & 73.02 & 11.66 & 21.12 & 48.74 & 38.48 \\
\midrule
CLIP                                         & \multirow{3}{*}{2-shot}  & 35.80 & 36.00 & 66.96 & 82.22 & 14.74 & 24.54 & 63.02 & 46.18 \\
UniCL                                        &                          & 35.60 & 40.00 & 60.86 & 85.40 & 16.54 & 22.34 & 60.60 & 45.91 \\
\rowcolor{Gray}Unimodal                      &                          & 36.50 & 43.00 & 70.46 & 84.70 & 14.52 & 23.98 & 65.00 & \textbf{48.31} \\
\rowcolor{Gray}DLILP                         &                          & 36.80 & 39.10 & 68.94 & 83.22 & 17.80 & 24.14 & 59.84 & 47.12 \\
\midrule
CLIP                                         & \multirow{3}{*}{4-shot}  & 40.50 & 39.30 & 64.40 & 88.70 & 18.10 & 23.68 & 70.76 & 49.35 \\
UniCL                                        &                          & 40.70 & 42.90 & 61.10 & 91.32 & 19.30 & 22.22 & 69.76 & 49.61 \\
\rowcolor{Gray}Unimodal                      &                          & 42.20 & 44.40 & 66.94 & 92.30 & 17.10 & 25.32 & 76.60 & \textbf{52.12} \\
\rowcolor{Gray}DLILP                         &                          & 42.80 & 43.30 & 69.54 & 90.22 & 19.86 & 23.66 & 70.52 & 51.41 \\
\midrule
CLIP                                         & \multirow{3}{*}{8-shot}  & 47.70 & 44.10 & 67.29 & 91.78 & 21.06 & 25.00 & 80.30 & 53.89 \\
UniCL                                        &                          & 45.90 & 47.50 & 63.60 & 92.76 & 20.78 & 25.60 & 77.74 & 53.41 \\
\rowcolor{Gray}Unimodal                      &                          & 50.20 & 48.50 & 66.60 & 93.58 & 21.14 & 28.26 & 80.80 & 55.58 \\
\rowcolor{Gray}DLILP                         &                          & 48.40 & 49.40 & 71.02 & 92.94 & 22.76 & 26.38 & 78.94 & \textbf{55.69} \\
\midrule
CLIP                                         & \multirow{3}{*}{16-shot} & 54.50 & 49.60 & 69.10 & 93.20 & 25.76 & 28.34 & 82.66 & 57.59 \\
UniCL                                        &                          & 53.10 & 50.90 & 65.58 & 93.78 & 24.56 & 26.94 & 80.38 & 56.46 \\
\rowcolor{Gray}Unimodal                      &                          & 54.20 & 53.70 & 67.68 & 94.36 & 26.20 & 30.26 & 81.62 & 58.29 \\
\rowcolor{Gray}DLILP                         &                          & 55.60 & 54.50 & 72.74 & 93.82 & 26.72 & 28.98 & 81.02 & \textbf{59.05}  \\
\bottomrule
\end{tabular}
\end{table}

\end{document}